\pdfoutput=1

\documentclass[11pt]{article}

\usepackage[final]{acl}

\usepackage{times}
\usepackage{latexsym}

\usepackage[T1]{fontenc}

\usepackage[utf8]{inputenc}

\usepackage{microtype}

\usepackage{inconsolata}

\usepackage{graphicx}
\usepackage{wrapfig}
\usepackage[inline]{enumitem}
\usepackage{multirow}
\usepackage{rotating}
\usepackage{amsmath}

\usepackage{booktabs}

\newcommand{\groupMetric}[1]{\textit{#1}}
\newcommand{\researchQuestion}[1][1]{\textit{RQ #1}}
\newcommand{\rdf}[1]{\texttt{\scriptsize{#1}}}

\usepackage{listings}
\usepackage{xcolor}

\definecolor{codegreen}{rgb}{0,0.6,0}
\definecolor{codegray}{rgb}{0.5,0.5,0.5}
\definecolor{codepurple}{rgb}{0.58,0,0.82}
\definecolor{backcolour}{rgb}{0.95,0.95,0.92}

\lstdefinestyle{mystyle}{
    backgroundcolor=\color{backcolour},   
    commentstyle=\color{codegreen},
    keywordstyle=\color{magenta},
    numberstyle=\color{codegray},
    stringstyle=\color{codepurple},
    basicstyle=\ttfamily\scriptsize,
    breakatwhitespace=false,         
    breaklines=true,                 
    captionpos=b,                    
    keepspaces=true,                 
    numbers=left,                    
    numbersep=5pt,                  
    showspaces=false,                
    showstringspaces=false,
    showtabs=false,                  
    tabsize=2
}
\lstdefinelanguage{MyDialogue}{
    alsodigit = {:},
    keywords = {Agent:, User:},
    morecomment=[l]{\#}
}
\usepackage{xcolor}

\definecolor{selcolor}{rgb}{0.53,0.30,1.00}
\definecolor{piekcolor}{rgb}{0.00,0.60,0.30}
\definecolor{todocolor}{rgb}{1.00,0.75,0.00}
\definecolor{deepercolor}{rgb}{0.42,0.27,0.57}
\definecolor{citationcolor}{rgb}{0.867,0.176,0.361}

\title{Knowledge acquisition for dialogue agents using reinforcement learning on graph representations}

\author{
 \textbf{Selene Baez Santamaria\textsuperscript{1}},
 \textbf{Shihan Wang\textsuperscript{2}},
 \textbf{Piek Vossen\textsuperscript{1}},
\\
\\
 \textsuperscript{1}Vrije Universiteit Amsterdam,
 \textsuperscript{2}Utrecht University
}

\begin{document}
\maketitle
\begin{abstract}
We develop an artificial agent motivated to augment its knowledge base beyond its initial training. The agent actively participates in dialogues with other agents, strategically acquiring new information. The agent models its knowledge as an RDF knowledge graph, integrating new beliefs acquired through conversation. Responses in dialogue are generated by identifying graph patterns around these new integrated beliefs. We show that policies can be learned using reinforcement learning to select effective graph patterns during an interaction, without relying on explicit user feedback. Within this context, our study is a proof of concept for leveraging users as effective sources of information. 
\end{abstract}

\section{Introduction}
\label{sec:intro}
Artificial interactive agents 
are designed to assist people. Usually, interaction modelling starts from the user's information need and not the system's information need. Such uni-directional modelling misses out 
to leverage the user as a knowledge source for the agent and not only as a knowledge seeker. To this end, we argue for \textbf{knowledge-centered} agents that can \begin{enumerate*}[label=(\roman*)]
    \item evaluate their knowledge state,
    \item evaluate their knowledge needs,
    \item acknowledge their lack of knowledge, and 
    \item actively try to obtain the missing knowledge through interaction with users.
\end{enumerate*} 

\begin{figure}[!ht]
    \centering
    \includegraphics[width=0.45\textwidth]{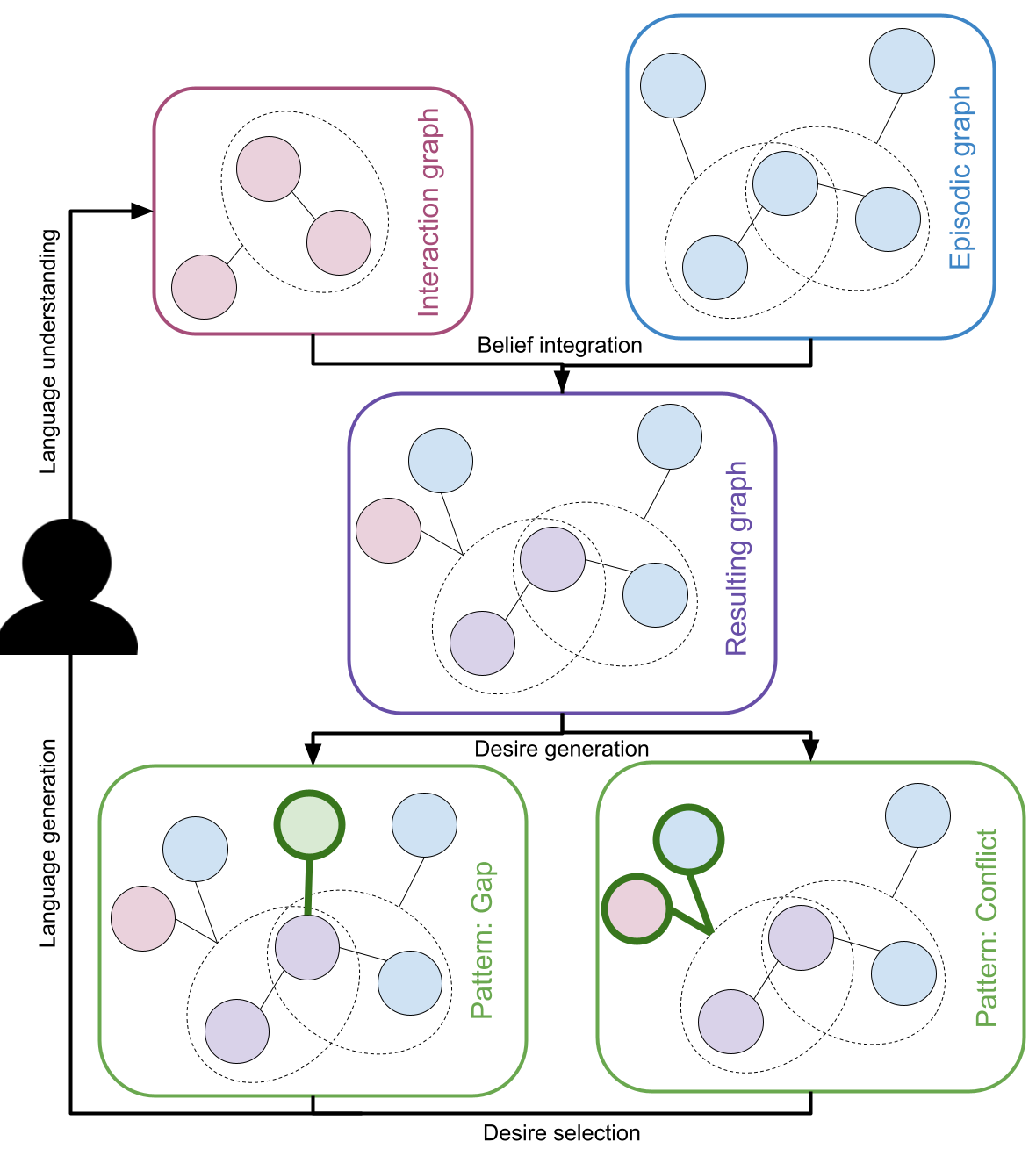}
    \caption{\footnotesize\label{fig:model}Dialogue management modelled as knowledge graphs. Information conveyed by the interlocutor at every turn is represented as triples in an interaction graph (in pink). This graph is integrated into the existing episodic knowledge graph of the artificial agent (in blue). We focus on specific graph patterns arising in the integrated neighbourhoods of the resulting graph (in purple). These might represent, for example, knowledge gaps (in green, bottom left) or conflicts (in green, bottom right). One pattern gets selected to respond to the interlocutor and continue the dialogue.}
\end{figure}

The knowledge targeted by such knowledge-centered agents might vary according to the application and shift during interactions. For some scenarios, an agent's goal may be to acquire in-depth knowledge on a given topic. For example, a customer service should know all factual information about the company's products, while a personal companion needs a complete overview of any relevant personal information to support a user. In contrast, for other scenarios, an agent should aim to gather diverse perspectives to break or expand self-imposed filter bubbles~\cite{aicher-etal-2022-towards-modelling}. For example, an online moderator should detect a wide range of opinions around the same topic~\cite{van2022hyena}, while news recommenders should provide complementary perspectives reporting events~\cite{reuver2021we}. Lastly, we argue that in any application, regardless of its training and performance, knowledge gaps may arise that need to be resolved and thus require active intervention of the agent. We therefore propose a solution to enhance agents with such generic capability.

In this paper, we present a knowledge-centered conversational agent that
\begin{enumerate}
\itemsep0em 
    \item Evaluates the status of its own knowledge.
    \item Can generate a wide range of responses in line with specific dialogue strategies to prompt the user to communicate further knowledge.
    \item Learns a dialogue policy to choose from these options in specific circumstances to improve its knowledge state.
\end{enumerate}

We provide evidence that artificial agents can drive conversation to pursue their own knowledge-centered goals by leveraging the user's knowledge, and without requiring explicit human feedback for learning. We formulate these goals at an abstract level that generalizes over specific application contexts and can therefore be used to adapt the agent's knowledge in  many applications. Hence, we step in the direction of developing conversational agents that become highly adaptable and responsive to a wide range of tasks and domains as they expand their knowledge. 

\section{Related work}
Knowledge-based conversational agents are an active area of research~\cite{ni2023recent}. Some approaches consider dialogue as a series of short  Q\&A tasks, where the usage of structured knowledge sources for retrieval of factual information particularly strengthens this type of dialogue~\cite{kim-etal-2023-task}. Another line of research adds a conversational layer to factual knowledge bases to facilitate querying them over natural language~\cite{ait2020kbot}. These techniques, however, fall short when a dialogue involves personal or opinion-based knowledge. 

Dialogue policy learning, particularly through reinforcement learning (RL), also shows substantial attention. Many studies address on Task-oriented Dialogue~\cite{rohmatillah2023advances} or Open-domain Dialogue~\cite{xu-etal-2020-conversational}. Few focus on the acquisition of knowledge, and these typically involve inquiring the user about their satisfaction with the interaction. However, in this work is concerned with filling domain or task-related knowledge gaps. For a similar approach, ~\citeauthor{mazumder2020continuous} propose a method for continuous open-world knowledge base completion within a conversational setting. 

\section{Framework description}
\label{sec:problem}
We propose a framework, formulated as a Belief-Desire-Intention (BDI) model~\cite{bratman1987intention}, where artificial agents have informational intents. In our approach, we model these intentions using symbolic knowledge bases. Specifically, we choose graph and RDF\footnote{Resource Description Framework: \url{https://www.w3.org/RDF/}} technologies to model the knowledge that agents either have or aim to have.

To explain our approach, we use the running example of an agent that has the goal to "know more" (as further defined in Section \ref{sec:intent-graph}). However, the proposed framework works for any informational intent, as long as this intention is measurable in the proposed symbolic representation.

\subsection{Defining a BDI model with KGs}
\label{sec:intent-graph}

\paragraph{Beliefs} We begin by modelling the informational state of the agent as a belief network, specifically as a knowledge graph where entity nodes are connected via semantically meaningful edges
Since the beliefs originate from the user input, we represent these as \rdf{CLAIMS} made by the user. These \rdf{CLAIMS} are the basic knowledge units, represented as RDF statements with subject-predicate-object triples. Each of these statements is embedded in its own RDF named graph~\cite{carroll2005named}, thus allowing a triple to serve as a node in other RDF statements. This simple, yet powerful knowledge representation technique allows to express complex and nested meanings (see Table \ref{table:claim}), where "there is knowledge about things", and "there is further knowledge about the known things"
. Furthermore, to recognize that the knowledge an agent has is not necessarily absolute, but rather a perspective on the real world, each \rdf{CLAIM} is associated to a \rdf{PERSPECTIVE}, hosting the particular source's certainty, polarity, and sentiment values of that belief. Through this modelling, an agent to hold contradictory, uncertain or ambivalent beliefs from multiple sources.

\begin{table*}[!h]
\centering
\scriptsize{
\caption{
\label{table:claim} Example of an agent's belief network. Top part showcases how knowledge units can be combined to express more complex knowledge. Bottom part showcases the quality of each knowledge unit, with specific polarity and certainty values. }
\begin{tabular}{|llll|} 
\hline
\textbf{Subject} & \textbf{Predicate} & \textbf{Object} & \textbf{Named Graph} \\ \hline \hline
\rdf{lWorld:diana}    & \rdf{n2mu:live} & \rdf{lWorld:paris} & \rdf{lWorld:diana\_live\_paris} \\
\rdf{lWorld:diana\_live\_paris}    & \rdf{n2mu:duration} & \rdf{lWorld:fiveYears} & \rdf{lTalk:diana\_live\_paris\_duration\_fiveYears} \\
\hline
\rdf{lWorld:diana\_live\_paris}    & \rdf{grasp:hasAttribution} & \rdf{lTalk:diana\_live\_paris\_01} & \rdf{lTalk:Perspectives} \\
\rdf{lTalk:diana\_live\_paris\_01} & \rdf{rdf:value}    & \rdf{certainty:uncertain} &  \rdf{lTalk:Perspectives} \\
\rdf{lTalk:diana\_live\_paris\_01} & \rdf{rdf:value}    & \rdf{polarity:positive} &  \rdf{lTalk:Perspectives} \\
\hline
\end{tabular}}
\end{table*}

\paragraph{Intentions} This tractable definition of beliefs allows an agent to evaluate the quality of its own knowledge by measuring specific aspects of its belief network. As a consequence, the agent is also equipped with the ability to set a target for any of these aspects. We regard these targets as the agent's informational intention, that is, the intended informational state of the agent. As a concrete example, an agent with the intention of having more complete knowledge (as introduced in Section \ref{sec:intro}) can be operationalized as an increasing volume of \rdf{CLAIMS}; while an agent with the intention of having more diverse knowledge can be operationalized as a growing volume of \rdf{PERSPECTIVES}. As such, any informational intention can be addressed under this framework, provided that the associated knowledge aspect can be measured on its belief network.

\paragraph{Desires} As the informational state of an agent changes, different graph patterns arise on its belief network. Specific graph patterns are semantically meaningful and are connected to different knowledge quality aspects, for example conflicting knowledge or novel knowledge. An agent can select any of these patterns to transform its current informational state into an intended one. Thus, we regard these semantic patterns as the desires of the agent that represent specific knowledge objectives relevant to its current informational state. In this paper we define eight abstract desires, as show in Figure \ref{fig:patterns-graphs}, each related to a specific knowledge aspect: correctness, completeness, redundancy, and interconnectedness~\cite{stvilia2007framework}.\footnote{This is not a comprehensive specification of patterns. Others could focus on \textit{complexity}, \textit{consistency}, or \textit{temporality} of knowledge.}  

\subsection{Knowledge acquisition modelled as KGs} 
\label{sec:interaction-graph}
So far we have focused on modelling an agent that can keep track of its current and intended informational state. Yet, we have not explained the mechanisms by which the agent acquires knowledge to transform that informational state. For this, an agent must engage in information-seeking behaviours~\cite{belkin1993interaction} and actively interact with sources in order to find the target knowledge. Similar to an information retrieval setting, two major features in the search of information are a) the modes of interaction, and b) the types of sources available. These two are typically intertwined, for instance, an interaction mode like "sensory experience" implies visual and auditory sources while a web search interaction mode implies sources like online textual news or Semantic Web databases like Wikidata. In this paper we experiment primarily with dialogue as an interaction mode, and human interlocutors as knowledge sources.


To be able to perform a dialogue with human interlocutors, our BDI network architecture needs to be integrated in a conversational agent. Throughout a conversation between an artificial agent and a knowledge source, we model the flow of information during their communication as episodic Knowledge Graphs ($eKG$), where each incoming utterance is transformed into RDF triples, and the accumulation of conversations is stored in a triple store~\cite{baez-santamaria-etal-2021-emissor}. For this purpose, an $eKG$ is conformed of five sub-graphs: \begin{enumerate*}[label=(\roman*)]
    \item \texttt{Ontology}: containing the world model,
    \item \texttt{Instances}: containing the individual entities in claims and their inter-claim connections,
    \item \texttt{Claims}: containing the set of atomic pieces of knowledge collected thus far,
    \item \texttt{Perspectives}: containing the specific viewpoint of the source regarding a claim,
    \item \texttt{Interactions}: containing the conversational provenance of each claim (e.g. source, place, and time of a chat). 
\end{enumerate*}

In addition to the above knowledge structure, the agent needs to be equipped with:
\begin{enumerate}
\itemsep-.5em 
\item \underline{language understanding} to interpret the interlocutor's input signal (e.g. audio, text, gestures) as providing an interaction knowledge graph ($iKG$, pink in Figure \ref{fig:model}),
\item \underline{belief integration} to merge the incoming beliefs ($iKG$) with the existing ones accumulated in the episodic knowledge graph ($eKG$, blue in Figure \ref{fig:model}),
\item \underline{desire generation} to evaluate the merged beliefs and produce a set of focused areas in the belief network to potentially improve upon on (green in Figure \ref{fig:model}), 
\item \underline{desire selection} to pick a specific belief that is to be changed by evoking the next interlocutor's input signal,
\item \underline{language generation} to formulate a response of the appropriate signal type (e.g. audio, text, gestures) to evoke the interlocutor response
\end{enumerate}

Through these five pipeline processes, the agent can create (pro-active) responses during conversation, where the BDI framework replaces a classical dialogue management module. The language understanding and language generation modules correspond to well-established NLU and NLG tasks. In this paper, we take the NLU and NLG components for granted and leave these for future work, as we are focusing here on the BDI graph framework.

\subsection{Measuring intent satisfaction by comparing KGs} 
\label{sec:reward}
As mentioned in Section \ref{sec:intent-graph}, intents are associated with the comparison of a current knowledge state and an intended knowledge state. To change its current knowledge state, an agent makes use of desires, one per time step, to gradually change its current knowledge state. As intents are associated to specific aspects that can be measured on an agent's belief network, it then follows that every desire can be evaluated in the following manner:
\begin{enumerate}
\itemsep-0.5em 
    \item Apply the intent-related metric $m$ on the agent's belief network $eKG$ at time $\tau$
    \item Select desire $d$ and use it in a information-seeking interaction (in this case, dialogue)
    \item Apply the intent-related metric $m$ on the agent's belief network $eKG$ at time $\tau + 1$
    \item Calculate the difference $\Delta m$ between the values of the intent-related metric before and after the desire $d$ was applied
    \item Determine whether the measured difference $\Delta m$ in the belief network contributes, hinders, or has no effect towards the intent
\end{enumerate}

\begin{figure}[!ht]
    \centering
    \includegraphics[width=0.5\textwidth]{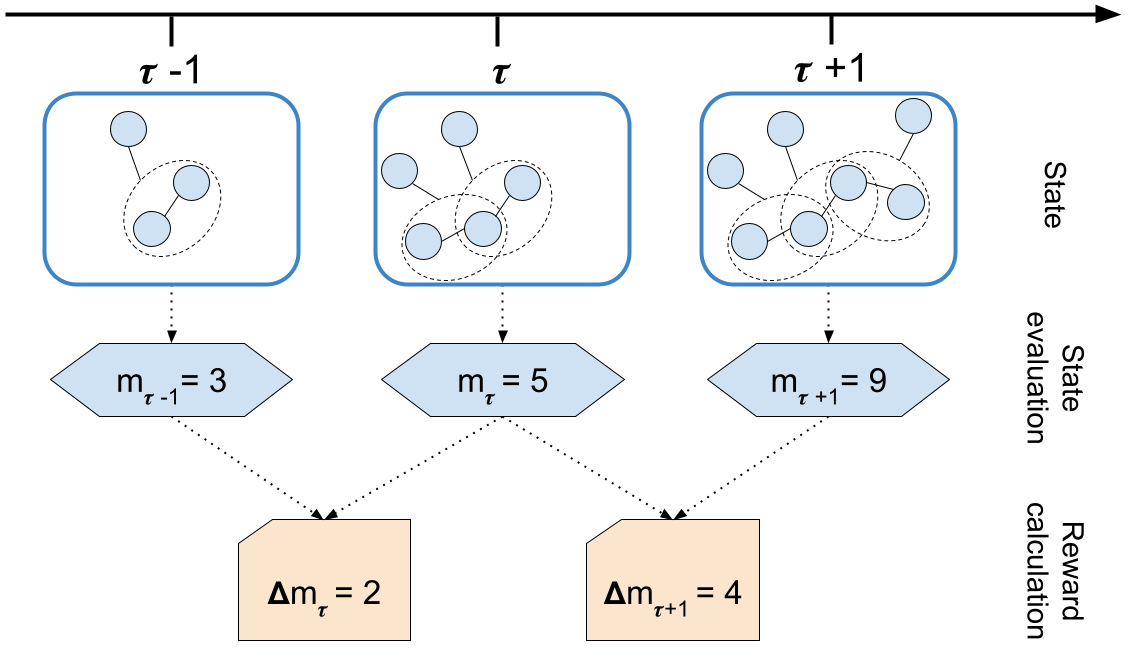}
    \caption{\footnotesize\label{fig:reward}Time-wise comparison of a belief network. At each time-step $\tau$, the knowledge state is assessed by applying metric $m$ on the $eKG$. In order to quantitatively evaluate the effect that a desire selected at time $\tau$ has on the belief network, the difference $\Delta m$ is calculated between the states $\tau -1$ and $\tau$.}
\end{figure}

Depending on the specific metric in question, the measured difference can vary in magnitude and direction. For the intention of having complete knowledge, operationalized as the metric of volume of \rdf{CLAIMS}, a positive difference \underline{contributes} to the intention as it signals that more \rdf{CLAIMS} have been added to the belief network, while a difference of $0$ \underline{has no effect} signals that, even if there had been changes to the belief network, these are not reflecting progress towards the intention.

This framework thus allows an agent, not only to have intentions and produce desires that pave a path towards satisfying this intention, but also provides a way to evaluate each desire's specific value, in the context of a given intention. 

\section{Methodology}
\label{sec:method}

The selection of desires is a crucial step in knowledge acquisition through dialogue. Thus, testing the utility of the proposed framework requires a method that learns which graph pattern (desire) will lead to the most valuable information (intent) in a specific but non-restrictive context.

For this, we use reinforcement learning (RL) to learn a policy that improves the relevance of the system's responses and augments the agent's learning abilities. We consider a fully observable environment where the state is the agent's accumulated $eKG$. The reward $r$ is calculated based on the comparison of consecutive states, as measured by a specific intent-related metric $m$. The problem presents a discrete action space, where the actions refer to the instantiated graph patterns $d$ and change with every interaction due to the specific entity and predicate types involved in the conversations. We aim to learn an optimal policy to determine which graph pattern to select. 

\begin{figure*}[!h]
    \centering
    \includegraphics[width=0.95\textwidth]{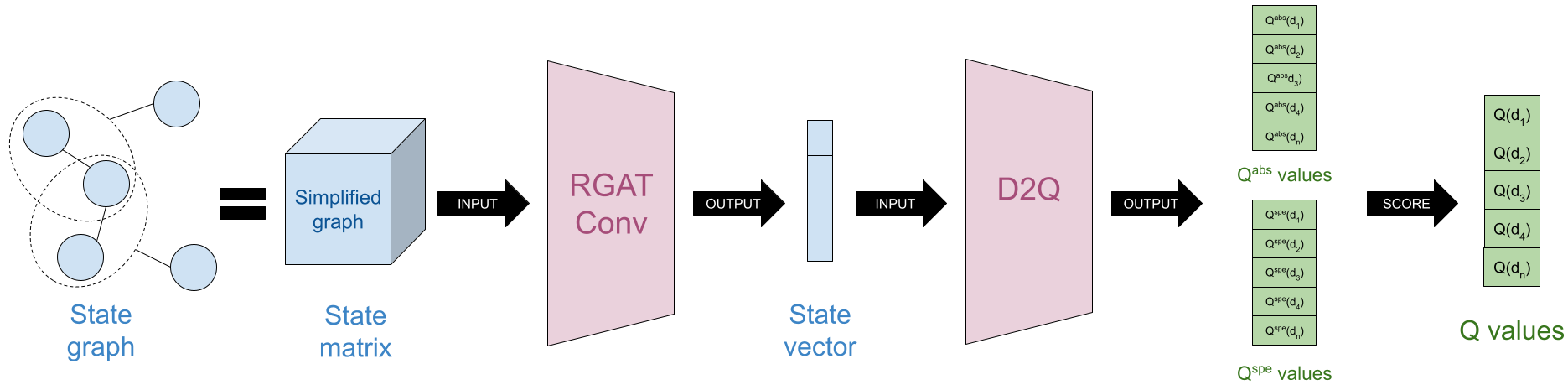}
    \caption{\footnotesize\label{fig:rl-dqn}Computational pipeline to calculate Q-values for different knowledge desires.}
\end{figure*}

\subsection{Problem formalization}
We formalize our RL problem as a discrete finite Markov decision process (MDP) and introduce the key components in the MDP as follows. 
\paragraph{State} The state is represented as Directed Acyclic Graph (DAG), specifically using the semantics of an $eKG$. This is formally defined as a tuple $eKG = (\mathcal{V}_e, \mathcal{E}_e, \mathcal{\varsigma}_e)$, where $\mathcal{V}$ is a set of nodes, $\mathcal{E}$ is a set of directed edges connecting pairs of nodes, and $\mathcal{\varsigma}$ is a set of statements. A statement is comprised of $\sigma = (s, p, o, c)$, where $s$ and $o\in\mathcal{V}$ are the subject and object entities, $p\in\mathcal{E}$ is the connecting relation and $c\in\mathcal{V}$ is the host named graph\footnote{Note that named graphs serve the function of encapsulating a single SPO triple that can later on be referred to in other statements, thus forming nested statements. As such, named graphs are both graphs $c\in\mathcal{C}$, and nodes themselves that can be head/tail entities in statements, resulting in $c\in\mathcal{V}$.}. Furthermore, $\mathcal{T}$ is a set of entity types and $\mathcal{P}$ is a set of predicate types. Every node has at least one entity type $t\in\mathcal{T}$ while every edge has exactly one predicate type $\rho\in\mathcal{R}$.

\paragraph{Action} Actions are generated by performing queries against the $eKG$, using information from the last $iKG$. As queries can also be represented as DAGs, each action type is also defined by a tuple of the form $d = (\mathcal{V}_a, \mathcal{E}_a, \mathcal{\varsigma}_a)$. The action space is defined by eight abstract graph query patterns, where each query pattern is characterized by a specific set of statements $\mathcal{\varsigma}_a$ containing either constant, instantiated or variable statement elements (full patterns are available on the Appendix, Table \ref{table:patterns}). As with any graph query, constant elements provide the semantics behind each action, while variable elements allow to search for a pattern in a given $eKG$. In contrast, instantiated elements are specific to the $iKG$ and modify an abstract query $d^{abs}$ on every dialogue turn thus making the actions $d^{spe}$ applicable to the current state transition. 

A selected action is sent in dialogue to the user, whose response generates an $iKG$ to be integrated to the agent's belief network. 

\paragraph{Transition} Given an $eKG$ at time $\tau$, it transitions to a new state at time $\tau + 1$ by incorporating an $iKG$ defined by a tuple $iKG = (\mathcal{V}_i, \mathcal{E}_i, \mathcal{\varsigma}_i)$. As mentioned before, an $iKG$ represents the content of an utterance by the user in dialogue, as shown in Table \ref{table:ikg}. Therefore, the structure of the $iKG$ is fixed by this specific set of statements $\mathcal{\varsigma}_i$, while the semantics are determined by the user and are reflected by instantiating $\mathcal{V}_i$, $\mathcal{E}_i$. 

At time $\tau$, there is no pre-established relation between the $eKG$ and its $iKG$. However, as the $iKG$ gets incorporated into the $eKG$ at time $\tau + 1$, then we can say that $\mathcal{V}_{i_\tau} \in \mathcal{V}_{e_{\tau+1}}$ and $\mathcal{E}_{i_\tau} \in \mathcal{E}_{e_{\tau+1}}$.

\paragraph{Reward function} As stated in Section \ref{sec:reward}, comparing two consecutive states allows to quantify the relative change in the belief network caused by selecting and employing the latest knowledge desire. We thus define reward $r$ as:

\begin{equation}
\mathbf{r}_\tau =\frac{ f(\mathcal{V}_e, \mathcal{E}_e, \mathcal{\varsigma}_e)_\tau }{ f(\mathcal{V}_e, \mathcal{E}_e, \mathcal{\varsigma}_e)_{\tau+1} } - 1
\label{eq:rgcn}
\end{equation}

For this, we require a metric $m$ to be applied to the belief network at each time step $\tau$. As mentioned in Section \ref{sec:intent-graph}, these metrics $m$ play the role of operationalizing a knowledge intent. 

\subsection{Policy optimization}
We optimize the policy $\pi$ to maps a state $eKG$ to an action $d$ (i.e. selecting the best graph pattern for a current $eKG$). Figure \ref{fig:rl-dqn} illustrates the architecture of this learning procedure. 

\paragraph{Representing the state} Given the complexity of the $eKG$, we create a simplified graph where the claims are the main nodes, connected to their respective perspective values. For this we extract the \texttt{Instances}, \texttt{Claims} and \texttt{Perspectives} subgraphs (described in  Section \ref{sec:interaction-graph}). This new simplified graph is centered around the perspective nodes, and their connection to claims thus represents the quality of what is known.

As node features we use the instances that are involved in the claims, using a one-hot-encoding representation. For the state encoder, we use an architecture with two RGAT layers~\cite{busbridge2019relational} followed by a fully connected layer, which results in node embeddings. To obtain a graph embedding we aggregate these via a mean operator. 

\paragraph{RL algorithm} We employ the D2Q algorithm~\cite{zhao2024decomposed} which provides a structure to separate \textit{abstract} actions from \textit{specific} actions thus mapping to our set of abstract and specific graph patterns. We consider abstract actions as the type of graph pattern to select (e.g. negation conflict) while the specific actions relate to the predicates and entities involved (e.g. conflict about \rdf{diana} \rdf{live} \rdf{paris}). Learning can be efficient by using the entity types (e.g. \rdf{person}, \rdf{city}) instead of the specific instances, allowing the agent to learn an approximation of a pattern's utility from fewer interactions.

The state vector is fed into a two-layers DQN architecture~\cite{mnih2013playing} to estimate the Q-values per action (hidden layer size = 64, replay memory size = 500). The output of this is fed into two parallel flows, each consisting of a fully connected layer and a final softmax layer. On the one hand, abstract actions are represented as the 8 possible graph patterns to choose from. On the other hand, specific actions are represented as all entity types available in a given ontology. 

Selecting an action consists of two steps: selecting an abstract action, and scoring the specific subactions. The abstract action is selected by taking the item with the highest value from its softmax head. For specific actions, a score is constructed as the weighted average of its entities types \(e\), using the values returned by the corresponding softmax head.  This constructive scoring method allows to score actions with novel combinations of entities.

\begin{figure*}[!h]
    \centering
    \includegraphics[width=.9\textwidth]{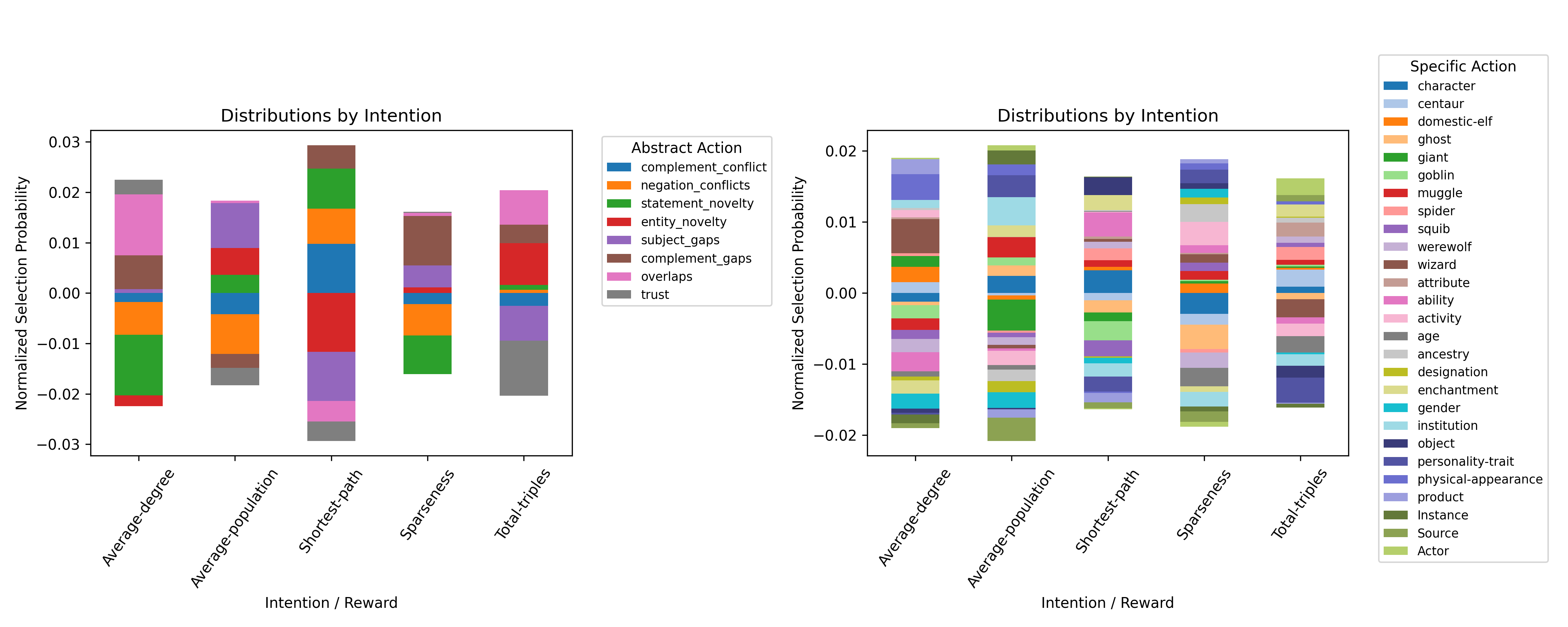}
    \caption{\footnotesize\label{fig:result1}Q-value distribution per abstract and specific action types (thought types and entity types respectively). Seed state is an empty graph, representing an agent with no knowledge on the topic yet. For visualization, the probabilities are normalized by subtracting the average probability. Shorter bars signal more equally distributed action values.}
\end{figure*}

\section{Experimental design}
We investigate the following research questions:

\begin{enumerate}[label=RQ\arabic*:] 
\itemsep-0.5em 
    \item\textbf{Characterizing agent behaviour} Do different agent intentions produce different dialogue strategies?  
    \item\textbf{Characterizing agent's knowledge} Do different agent intentions acquire different knowledge? 
    \item\textbf{Impact of the source} How do different knowledge sources impact the learning process of agents with different intentions? 
\end{enumerate}

\subsection{Experimental conditions}
We investigate 8 knowledge intents, operationalized with the graph metrics described in Table \ref{table:rewards}. As different metrics measure distinctive aspects of knowledge, we hypothesize that each metric will produce distinct agent behaviours.

\begin{table}[!h]
\centering
\scriptsize{
\caption{
\label{table:rewards}Graph metrics and their knowledge-centered intention. Knowledge dimensions are inspired from~\cite{nurse2011information}}
\begin{tabular}{|p{1.55cm}|p{1.7cm}|c|}
\hline
\textbf{Metric}     & \textbf{Dimension}  & \textbf{Formula} \\ \hline \hline
Sparseness	& Cohesion & $\frac{\mathcal{E}}{\mathcal{V}^2}$ \\
Average degree	& Interconnectedness & $\frac{2 |\mathcal{E} |}{\mathcal{V}}$ \\
Shortest path	& Specificity & $\frac{1}{\mathcal{V}} \sum_{i \neq j} min(dist(v_i, v_j)) $  \\
Total triples	& Volume & $\mathcal{\varsigma}$  \\
Average population	& Spread & $\frac{1}{\mathcal{T}}\sum_{t \in \mathcal{T}} |e_t| $ \\
Ratio claims to triples	& Completeness & $\frac{|{claims}|}{|\mathcal{\varsigma}|}$ \\
Ratio perspectives to claims	& Diversity & $\frac{|{perspectives}|}{|{claims}|}$ \\
Ratio conflicts to claims	& Correctness& $\frac{|{conflicts}|}{|{claims}|}$ \\ \hline
\end{tabular}}
\end{table}

We setup two experiments. In the first, the knowledge-centered agents converse with a single user with perfect knowledge. In the second experiment, the agents are exposed to users with varying knowledge quality to simulate the diversity of knowledge sources available in the wild.

\subsection{Evaluation}
\label{sec:evaluation}
To answer \researchQuestion[1], we compare the dialogue policies learned by agents with different intentions/rewards. This is estimated by the Q-values produced by the D2Q network, as these indicate the expected return (associated with the reward) by taking different actions given a certain state. Since the Q-values are state dependent, we take as use case an empty $eKG$, representing the beginning of a conversation and when the tone and topic are established.

To answer \researchQuestion[2], we compare the belief networks of agents with different intentions/rewards. This is performed by measuring its knowledge interconnectedness, specificity, volume, spread, completeness, diversity and correctness as operationalized by the 8 metrics previously selected as rewards. 

To answer \researchQuestion[3], we analyze the changes  in the rewards obtained by agents conversing with users with perfect knowledge vs the ones exposed to users with imperfect knowledge.

\subsection{Data}
We utilize the Harry Potter Dialogue (HPD) dataset~\cite{chen-etal-2023-large} which also contains structured information about characters in the novel. 
Furthermore, the data is temporally divided according to seven books, thus allowing to simulate conversations over time where some attributes change, while others remain stable. We transform the data into RDF triples, removing invalid punctuation and splitting lists into individual values. The dataset characteristics are shown in Table \ref{table:hpd}.

\subsection{User model}
Five user model types are created as knowledge bases of varied quality (Table \ref{table:users}). To simulate a conversation, the selected graph pattern $d$ is transformed into a SPARQL query that can be run against the user model's triple store. The response triples are formatted as an $iKG$, representing the acquired knowledge from the user. Please note that not all possible graph patterns $d$ will result in a successful query to the user model in which case, the user model will randomly select a piece of knowledge, as a way to continue the dialogue. 

\subsection{Training setup}
\label{sec:train}




Dialogue is carried out in RDF form directly to isolate the dialogue policy optimization. As such, we do not include speech detection or generation. Similarly, information extraction to transform natural language intro RDF triples, and Natural Language Generation fall out of scope. Therefore, the optimization focuses on learning policies for choosing adequate graph patterns and is not influenced by errors from other pipeline systems. The agents are trained for 8 conversations of 20 turns each (10 for the human and 10 for the agent). We perform an update on the policy on every agent turn, resulting in 80 (10x8) policy updates. As the graphs get reset every second conversation, the maximum number of state transitions is 20 (10x2). The network is saved at the end of every conversation, resulting in 8 checkpoints. We run each setting 3 times and present the average results.
More details about training mechanisms and parameter settings in the RL algorithm are presented in Appendix \ref{training details}.

\section{Results and discussion}
\label{sec:results}
\begin{figure}[!h]
    \centering
    \includegraphics[width=0.45\textwidth]{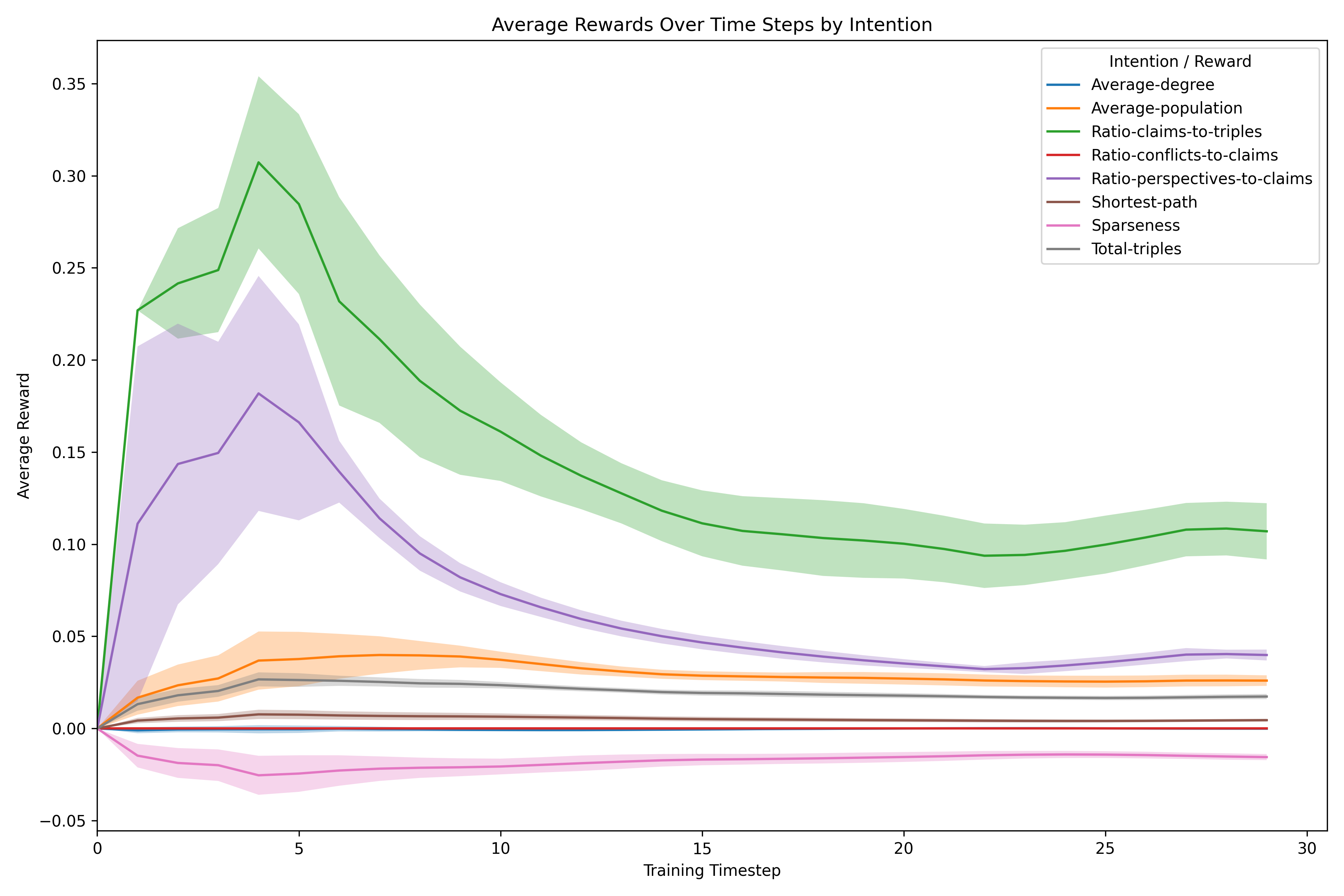}
    \caption{\footnotesize\label{fig:resultt1}Average rewards obtained by the policy at every training step.}
\end{figure}

We first evaluate the training process per intention, by calculating the average rewards during training under the corresponding reward function. In Figure~\ref{fig:resultt1}, we observe that 5 metrics stabilize in their learning, while 3 of them do not. Taking the learning curve of Average-population (in orange) as an example, the average reward increases during the early timesteps and converges towards a stable level. This phenomenon shows the early learning process of the RL algorithm and indicates its capability of finding a stable policy that can select the best graph pattern for an $eKG$ under this intention. 
Looking back at Table~\ref{table:rewards}, those metrics that learn well are defined based on structural aspects of the graph, while those defined as semantic ratios have difficulty guiding the RL algorithm. This might signal that semantic rations have more complex correlations (or maybe causal relations) between the number of claims and the number of turns or consecutive conversations.  

From this point forward we focus our analysis on the 5 intentions that proved fast learning convergence into stable learned policies.

\paragraph{Learned dialogue policies (RQ1)} 
Figure \ref{fig:result1} shows the distribution of action values per intention of the learned policies where some intentions are more equally distributed, like Sparseness, while others have a wider probability range like Shortest-path. We note that some abstract actions are consistently preferred, like Overlaps, while other abstract actions are mostly excluded, like Trust. Regardless of the overall trends, we can confirm that different intentions produce distinct dialogue strategies. 

For example, Average degree can be characterized by dialogues where known information is mentioned in order to get the user's perspective (Agent: "Did you know that Ginny has red hair, just like Ron?", User: "No, I am sure that she does not have red hair") combined with trust judgments towards the user, based these perspectives (Agent: "I do not trust you"). This type of policy implicitly improves the interconnectedness between what is known and the user perspectives on this knowledge, thus profiling the knowledge source.

While Average population and Total triples also prompt the user for their perspective on what is known, in contrast these combine it with further questions regarding subjects (Agent: "What color is Ginny's hair") or objects (Agent: "Who has red hair then?") respectively. Interestingly enough, these two policies actively avoid making trust judgments on the user, and instead focus on expanding their knowledge base further. 

\begin{figure}[!h]
    \centering
    \includegraphics[width=0.45\textwidth]{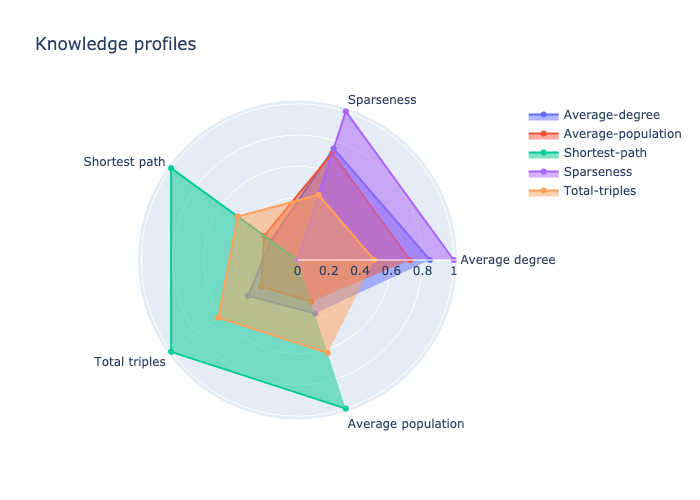}
    \caption{\footnotesize\label{fig:result2}Profiles of the knowledge acquired by different intentions. Knowledge dimensions are operationalized according to the graph metrics in Table~\ref{table:rewards}.}
\end{figure}

\paragraph{Acquired knowledge (RQ 2)} 
We analyze the final $eKG$ according to the 5 aforementioned metrics (Figure~\ref{fig:result2}, further details on Table~\ref{table:final_knowledge}). Overall, we see evidence that three specific knowledge profiles arise, distinguished by different intentions. The intentions Sparseness, Average degree and Average population generate similar knowledge profiles more centered around knowledge cohesion and interconnectedness. Shortest path as an intention focuses more on the volume, spread and specificity of knowledge. Total triples instead keep a balanced profile, keeping most of the knowledge aspects at an equal level.

\begin{figure}[!h]
    \centering
    \includegraphics[width=0.45\textwidth]{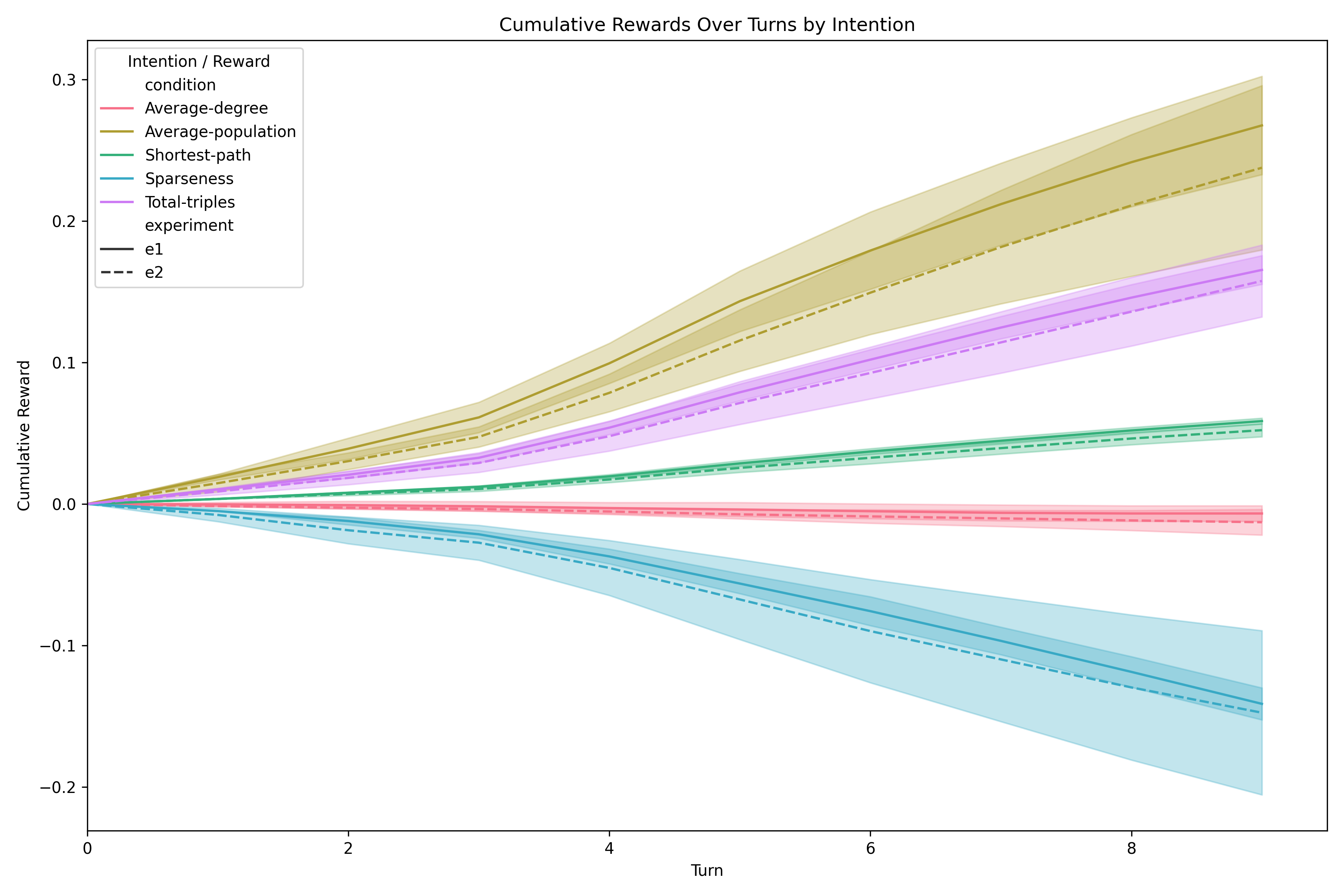}
    \caption{\footnotesize\label{fig:result3}Cumulative rewards per intention using the trained network. Comparison between experiment 1 (perfect knowledge user) and experiment 2 (imperfect knowledge user).}
\end{figure}

\paragraph{Policy updates (RQ 3)} 
We investigate the effects of imperfect knowledge sources by comparing the cumulative reward for each intention across experiments 1 (user model with perfect knowledge) and experiment 2 (user models with imperfect knowledge). Figure \ref{fig:result3} shows rewards are consistently lower when the agents are exposed to imperfect knowledge sources, however, some rewards (e.g. Average population) are more sensitive than others (e.g. Average degree). This can be explained by looking back at the learned dialogue policies analyzed in RQ1. While trying to expand its knowledge, Average population poses more questions to the user, which can lead to unanswered questions given an imperfect knowledge source. In contrast, Average degree focuses on profiling the knowledge source itself, which can be done regardless of the quality of the knowledge source.

\section{Conclusion}
In this work we propose a theoretical and mathematical framework for conversational agents to pursue their own knowledge goals in open-domain settings. In this framework, specific knowledge goals (or intentions) can be operationalized as domain independent graph metrics. We provide evidence that some graph metrics can quickly learn stable and optimal dialogue policies via reinforcement learning, and analyze such resulting dialogue policies. We test these dialogue policies and compare the knowledge gathered by each of them. Finally, we demonstrate that this framework is robust to knowledge sources of different quality.

\newpage
\section*{Limitations}
In this work we use operationalize knowledge quality aspects as measurable graph properties. Though this has been proposed carefully, the terminology might be too coarse for other specialized disciplines like epistemology.

On a different aspect, the scalability of the proposed methods are to be further examined. As there are no restrictions on the size or structure of the $eKG$, the state space is infinite and the learning procedure can be challenging when the state space gets too big.

\section*{Ethics statement}
The framework proposed in this study aims to enable artificial agents to pursue knowledge driven goals, utilizing people as knowledge sources. Depending on the application and the users available, the misuse of these technologies might result in concerns about privacy and monitoring, particularly with vulnerable groups.



\bibliography{anthology,custom}

\begin{thebibliography}{19}
\providecommand{\natexlab}[1]{#1}

\bibitem[{Aicher et~al.(2022)Aicher, Minker, and Ultes}]{aicher-etal-2022-towards-modelling}
Annalena Aicher, Wolfgang Minker, and Stefan Ultes. 2022.
\newblock \href {https://aclanthology.org/2022.lrec-1.438} {Towards modelling self-imposed filter bubbles in argumentative dialogue systems}.
\newblock In \emph{Proceedings of the Thirteenth Language Resources and Evaluation Conference}, pages 4126--4134, Marseille, France. European Language Resources Association.

\bibitem[{Ait-Mlouk and Jiang(2020)}]{ait2020kbot}
Addi Ait-Mlouk and Lili Jiang. 2020.
\newblock Kbot: a knowledge graph based chatbot for natural language understanding over linked data.
\newblock \emph{IEEE Access}, 8:149220--149230.

\bibitem[{Baez~Santamaria et~al.(2021)Baez~Santamaria, Baier, Kim, Krause, Kruijt, and Vossen}]{baez-santamaria-etal-2021-emissor}
Selene Baez~Santamaria, Thomas Baier, Taewoon Kim, Lea Krause, Jaap Kruijt, and Piek Vossen. 2021.
\newblock \href {https://aclanthology.org/2021.mmsr-1.6} {{EMISSOR}: A platform for capturing multimodal interactions as episodic memories and interpretations with situated scenario-based ontological references}.
\newblock In \emph{Proceedings of the 1st Workshop on Multimodal Semantic Representations (MMSR)}, pages 56--77, Groningen, Netherlands (Online). Association for Computational Linguistics.

\bibitem[{Belkin et~al.()}]{belkin1993interaction}
Nicholas~J Belkin et~al.
\newblock Interaction with texts: Information retrieval as information seeking behavior.

\bibitem[{Bratman(1987)}]{bratman1987intention}
Michael Bratman. 1987.
\newblock Intention, plans, and practical reason.

\bibitem[{Busbridge et~al.(2019)Busbridge, Sherburn, Cavallo, and Hammerla}]{busbridge2019relational}
Dan Busbridge, Dane Sherburn, Pietro Cavallo, and Nils~Y Hammerla. 2019.
\newblock Relational graph attention networks.
\newblock \emph{arXiv preprint arXiv:1904.05811}.

\bibitem[{Carroll et~al.(2005)Carroll, Bizer, Hayes, and Stickler}]{carroll2005named}
Jeremy~J Carroll, Christian Bizer, Pat Hayes, and Patrick Stickler. 2005.
\newblock Named graphs.
\newblock \emph{Journal of Web Semantics}, 3(4):247--267.

\bibitem[{Chen et~al.(2023)Chen, Wang, Jiang, Cai, Li, Chen, Wang, and Li}]{chen-etal-2023-large}
Nuo Chen, Yan Wang, Haiyun Jiang, Deng Cai, Yuhan Li, Ziyang Chen, Longyue Wang, and Jia Li. 2023.
\newblock \href {https://doi.org/10.18653/v1/2023.findings-emnlp.570} {Large language models meet harry potter: A dataset for aligning dialogue agents with characters}.
\newblock In \emph{Findings of the Association for Computational Linguistics: EMNLP 2023}, pages 8506--8520, Singapore. Association for Computational Linguistics.

\bibitem[{Kim et~al.(2023)Kim, Gella, Zhao, Jin, Papangelis, Hedayatnia, Liu, and Z~Hakkani-Tur}]{kim-etal-2023-task}
Seokhwan Kim, Spandana Gella, Chao Zhao, Di~Jin, Alexandros Papangelis, Behnam Hedayatnia, Yang Liu, and Dilek Z~Hakkani-Tur. 2023.
\newblock \href {https://aclanthology.org/2023.dstc-1.29} {Task-oriented conversational modeling with subjective knowledge track in {DSTC}11}.
\newblock In \emph{Proceedings of The Eleventh Dialog System Technology Challenge}, pages 274--281, Prague, Czech Republic. Association for Computational Linguistics.

\bibitem[{Mazumder et~al.(2020)Mazumder, Liu, Ma, Wang, and Amazon}]{mazumder2020continuous}
Sahisnu Mazumder, Bing Liu, Nianzu Ma, Shuai Wang, and AI~Amazon. 2020.
\newblock Continuous and interactive factual knowledge learning in verification dialogues.
\newblock In \emph{NeurIPS-2020 Workshop on Human And Machine in-the-Loop Evaluation and Learning Strategies}.

\bibitem[{Mnih et~al.(2013)Mnih, Kavukcuoglu, Silver, Graves, Antonoglou, Wierstra, and Riedmiller}]{mnih2013playing}
Volodymyr Mnih, Koray Kavukcuoglu, David Silver, Alex Graves, Ioannis Antonoglou, Daan Wierstra, and Martin Riedmiller. 2013.
\newblock Playing atari with deep reinforcement learning.
\newblock \emph{arXiv preprint arXiv:1312.5602}.

\bibitem[{Ni et~al.(2023)Ni, Young, Pandelea, Xue, and Cambria}]{ni2023recent}
Jinjie Ni, Tom Young, Vlad Pandelea, Fuzhao Xue, and Erik Cambria. 2023.
\newblock Recent advances in deep learning based dialogue systems: A systematic survey.
\newblock \emph{Artificial intelligence review}, 56(4):3055--3155.

\bibitem[{Nurse et~al.(2011)Nurse, Rahman, Creese, Goldsmith, and Lamberts}]{nurse2011information}
Jason~RC Nurse, Syed~Sadiqur Rahman, Sadie Creese, Michael Goldsmith, and Koen Lamberts. 2011.
\newblock Information quality and trustworthiness: A topical state-of-the-art review.

\bibitem[{Reuver et~al.(2021)Reuver, Mattis, Sax, Verberne, Tintarev, Helberger, Moeller, Vrijenhoek, Fokkens, and van Atteveldt}]{reuver2021we}
Myrthe Reuver, Nicolas Mattis, Marijn Sax, Suzan Verberne, Nava Tintarev, Natali Helberger, Judith Moeller, Sanne Vrijenhoek, Antske Fokkens, and Wouter van Atteveldt. 2021.
\newblock Are we human, or are we users? the role of natural language processing in human-centric news recommenders that nudge users to diverse content.
\newblock In \emph{Proceedings of the 1st Workshop on NLP for Positive Impact}, pages 47--59.

\bibitem[{Rohmatillah et~al.(2023)Rohmatillah, Chien et~al.}]{rohmatillah2023advances}
Mahdin Rohmatillah, Jen-Tzung Chien, et~al. 2023.
\newblock Advances and challenges in multi-domain task-oriented dialogue policy optimization.
\newblock \emph{APSIPA Transactions on Signal and Information Processing}, 12(1).

\bibitem[{Stvilia et~al.(2007)Stvilia, Gasser, Twidale, and Smith}]{stvilia2007framework}
Besiki Stvilia, Les Gasser, Michael~B Twidale, and Linda~C Smith. 2007.
\newblock A framework for information quality assessment.
\newblock \emph{Journal of the American society for information science and technology}, 58(12):1720--1733.

\bibitem[{van~der Meer et~al.(2022)van~der Meer, Liscio, Jonker, Plaat, Vossen, and Murukannaiah}]{van2022hyena}
Michiel van~der Meer, Enrico Liscio, Catholijn~M Jonker, Aske Plaat, Piek Vossen, and Pradeep~K Murukannaiah. 2022.
\newblock Hyena: A hybrid method for extracting arguments from opinions.

\bibitem[{Xu et~al.(2020)Xu, Wang, Niu, Wu, Che, and Liu}]{xu-etal-2020-conversational}
Jun Xu, Haifeng Wang, Zheng-Yu Niu, Hua Wu, Wanxiang Che, and Ting Liu. 2020.
\newblock \href {https://doi.org/10.18653/v1/2020.acl-main.166} {Conversational graph grounded policy learning for open-domain conversation generation}.
\newblock In \emph{Proceedings of the 58th Annual Meeting of the Association for Computational Linguistics}, pages 1835--1845, Online. Association for Computational Linguistics.

\bibitem[{Zhao et~al.(2024)Zhao, Yin, Wang, Dastani, and Wang}]{zhao2024decomposed}
Yangyang Zhao, Kai Yin, Zhenyu Wang, Mehdi Dastani, and Shihan Wang. 2024.
\newblock Decomposed deep q-network for coherent task-oriented dialogue policy learning.
\newblock \emph{IEEE/ACM Transactions on Audio, Speech, and Language Processing}.

\end{thebibliography}

\newpage
\appendix

\section{Appendix}
\label{sec:appendix}

\subsection{Desires as abstract graph patterns}
\begin{figure*}[!ht]
    \centering
    \includegraphics[width=0.95\textwidth]{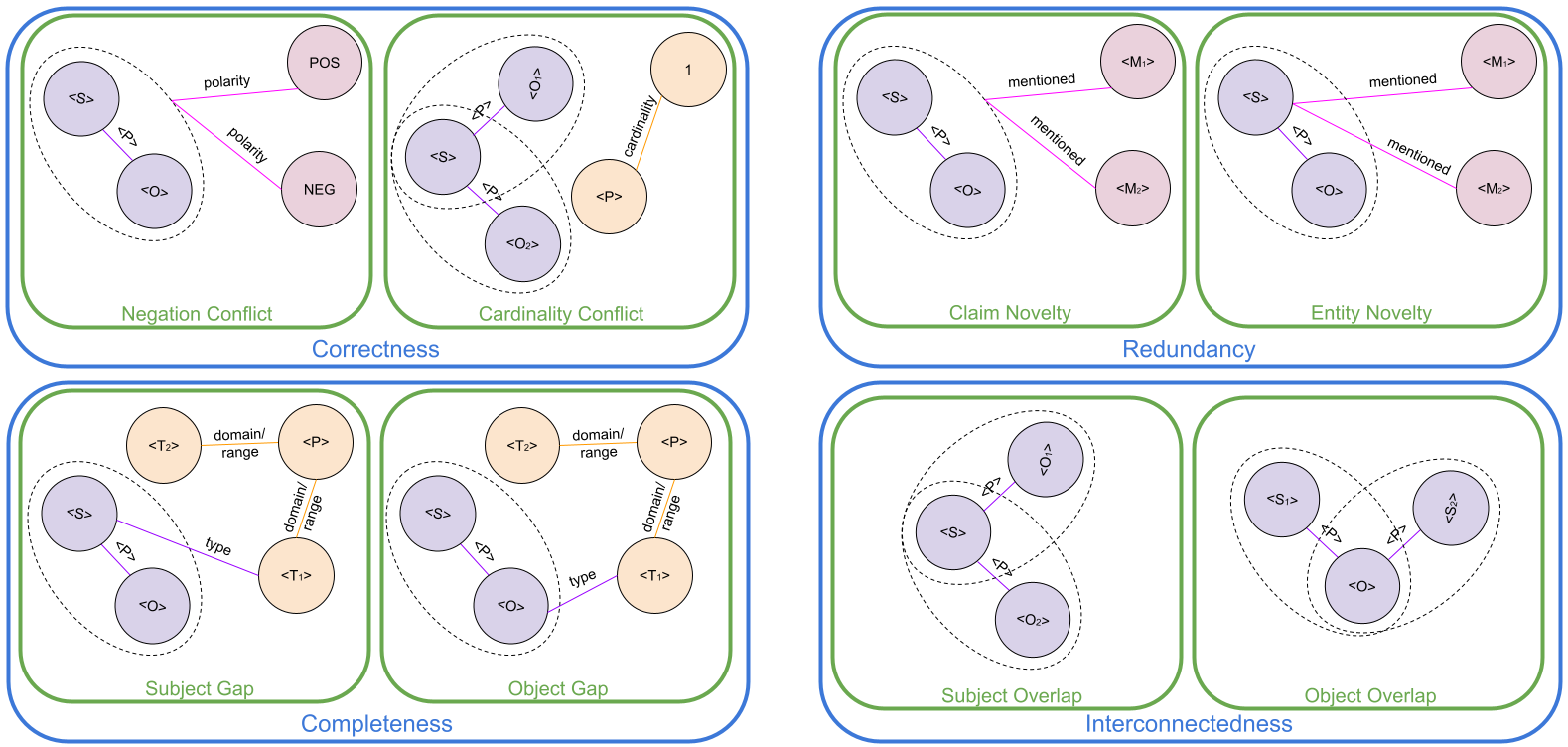}
    \caption{\label{fig:patterns-graphs}Simplified graphic visualization of semantic graph patterns that represent informational desires. Green boxes encapsulate specific graph structures, while blue boxes group graph structures associated to similar knowledge aspects. Nodes are represented as circles. Edges are represented as continuous lines between them. Named graphs are represented as dashed circles around a single triple. Elements in purple represent \rdf{CLAIMS}, elements in pink represent \rdf{PERSPECTIVES}, and elements in orange represent the \rdf{ONTOLOGY}. Items between <> represent elements to be instantiated in a specific belief network.}
\end{figure*}

We operationalize conversational desires (under the proposed BDI model) as abstract RDF graph patterns. There are specified as triple patterns in Table \ref{table:patterns} and visualized in Figure \ref{fig:patterns-graphs}.

\subsection{Dialogue management for knowledge acquisition}
The details of the dialogue management process as a BDI model are explained below:

\paragraph{Belief integration:} As input, the knowledge integration step takes a) an interaction knowledge graph ($iKG$) with factoids acquired in the last conversational turn and b) an episodic knowledge graph ($eKG$) containing the accumulated information acquired by the artificial agent thus far. Table \ref{table:ikg} illustrates how an $iKG$ represents the incoming beliefs and their provenance. An $eKG$ is a collection of $iKG$, thus following a similar but larger structure. 

\paragraph{Desire generation:} As explained in Section \ref{sec:intent-graph}, the current framework proposes eight tailored graph patterns that evaluate four different knowledge aspects: correctness, completeness, redundancy, and interconnectedness. Each of these abstract patterns can be instantiated with the specific Subject, Predicate and Object present in the $iKG$, which typically produces a wide range of specific desires. Thus, each of these desires targets a concrete belief that the agent intends to change in a particular knowledge quality direction. 

\paragraph{Desire selection:} A single desire is selected to form a response and continue the dialogue. Different system responses vary significantly in relevance and semantic plausibility, so they elicit distinct counter-responses from the human interlocutor. Therefore, the agent's chances of acquiring knowledge of sufficient quality highly depend on the selected desire.

\subsection{Dataset}
Here we show some statistics on the range and domain of the different predicates in the Harry Potter Dialogue~\cite{chen-etal-2023-large} dataset. This information might bring insight into which abstract thought patterns are better suited per predicate type. Predicates with a large domain scope (e.g Gender) are better paired with object gaps and object overlaps, while predicates with large range scope benefit from subject gaps and subject overlaps.


\begin{table}[h]
\centering
\small{
\caption{
\label{table:hpd}Dataset statistics, after converted to RDF.  For each role (Object or Subject) the number of distinct entities present in that role is reported. }
\begin{tabular}{|p{1.7cm}|c|c|}
\hline
\textbf{Predicate} & \textbf{Range (Object)}  & \textbf{Domain (Subject)}  \\ \hline \hline
Looks & 428 & 107 \\ \hline
Spells & 200 & 47 \\ \hline
Belongings & 189 & 49 \\ \hline
Title & 101 & 86 \\ \hline
Personality & 39  & 46 \\ \hline
Affiliation & 27 & 94 \\ \hline
Hobbies & 23 & 22 \\ \hline
Export & 16 & 24 \\ \hline
Talents & 15 & 13 \\ \hline
Lineage & 11 & 83 \\ \hline
Age & 11 & 106 \\ \hline
Gender & 2 & 124 \\ \hline
\end{tabular}}
\end{table}


\subsection{User models}
Five types of user models are used in this work, as described in Table \ref{table:users}. The first one is modelled with perfect knowledge while the other four types have imperfect knowledge. When creating each type, the base vanilla user is corrupted in a specific way, as described on the last column of the table. For each of the imperfect user types, 100 instances were generated.

\subsection{Training and parameters}
\label{training details}
In order to facilitate learning we introduce two training mechanisms: reset and shuffle. Reset, on the one hand, clears out the $eKG$ and restarts it to an empty condition. This mechanism counters the fact that since we measure the changes on the $eKG$ and the $eKG$ keeps growing, it may have a chance that the same action leads to different rewards when the $eKG$ is getting bigger. Shuffle, on the other hand, swaps the $eKG$ with another random one of similar size. This mechanism exposes the networks to more varied states thus prevents the networks from learning simply the specific state transitions. In the experiments, we reset the $eKG$ every 2 conversations and shuffle every 2 conversations in an alternating manner. The D2Q network is optimized with a learning rate of $1e-4$, a batch size of $4$, a $\gamma$ factor of $0.99$ and a $\tau$ value of $0.005$. The experiments were run on an NVIDIA A10 GPU for $~5$ hours.

\subsection{Extra results}
\begin{figure}[!h]
    \centering
    \includegraphics[width=.45\textwidth]{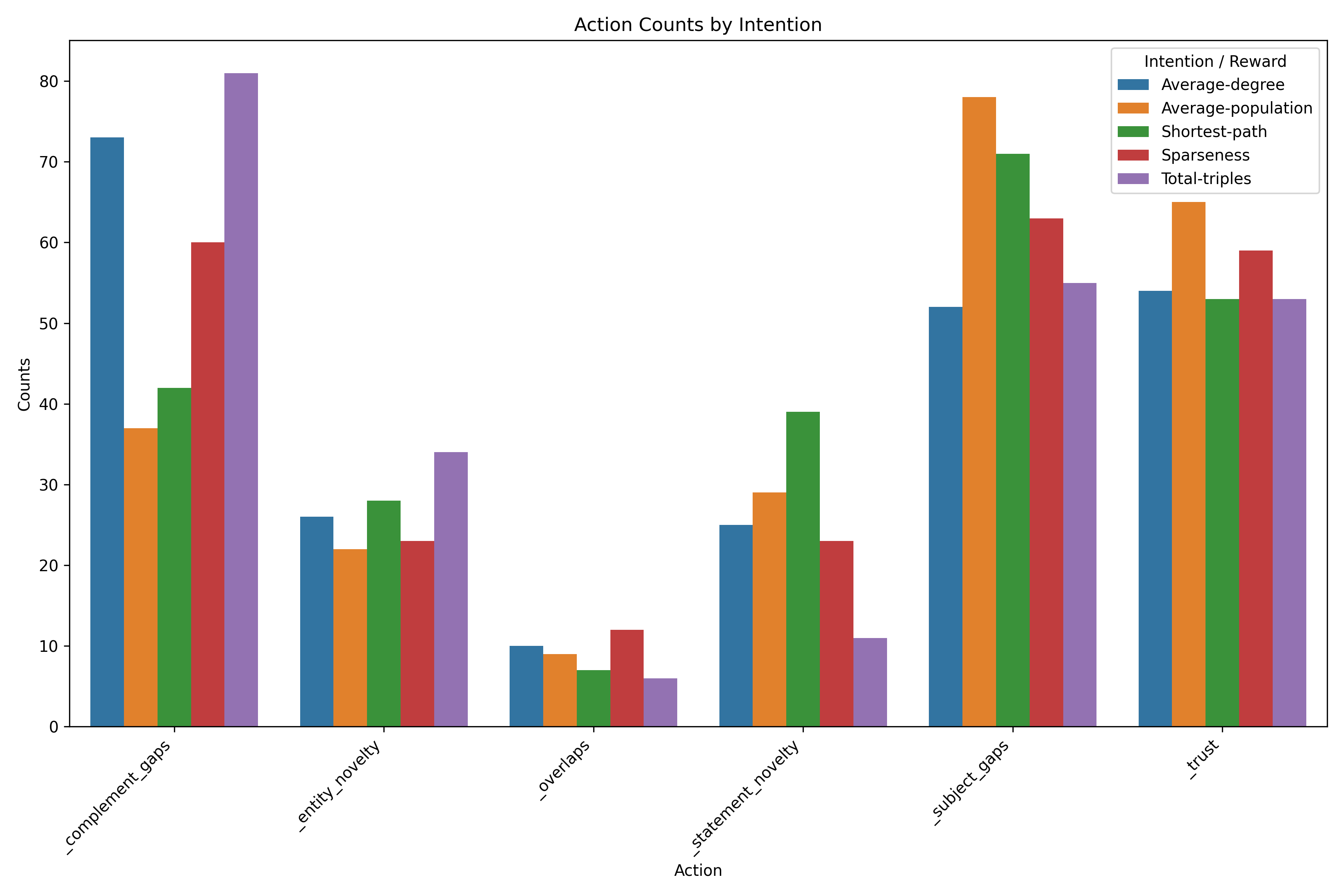}
    \caption{\label{fig:resultcounts}Abstract action counts selected during testing chat by the optimized trained network, averaged over runs.}
\end{figure}

Figure~\ref{fig:resultcounts} shows the selection counts per action per intention. This is further evidence for RQ 1 that distinct dialogue strategies arise. 

As further evidence for RQ 2, Table \ref{table:final_knowledge} reports the values for the 5 metrics on the final eKGs for different intentions.

\begin{table}[!h]
\centering
\scriptsize{
\caption{
\label{table:final_knowledge}Description of the knowledge acquired under different intentions, as measured by different graph metrics. Test run for one conversation of 10 turns, using the frozen optimized policy network.}
\begin{tabular}{|p{1.35cm}|p{.75cm}|p{.75cm}|p{.75cm}|p{.75cm}|p{.75cm}|}
\hline
Reward &  \textbf{Average degree} & \textbf{Sparseness} & \textbf{Shortest path} & \textbf{Total triples} & \textbf{Average population} \\ \hline \hline
    Average-degree &          12.377 &       0.745 &          2.555$^*$ &           4222$^*$ &              21.000$^*$ \\
Average-population &          12.406 &       0.756 &          2.548 &           4170 &              20.320 \\
     Shortest-path &          12.492$^*$ &       0.780$^*$ &          2.530 &           4076 &              19.033 \\
        Sparseness &          12.452 &       0.765 &          2.541 &           4146 &              19.974 \\
     Total-triples &          12.398 &       0.751 &          2.551 &           4197 &              20.680 \\
\hline
\end{tabular}}
\end{table}

\begin{table*}[h]
\centering
\small{
\caption{
\label{table:patterns}Semantic graph patterns. Items between <> represent variable nodes. \underline{<UNDERLINED>} items represent nodes that need to be instantiated.}
\begin{tabular}{|p{1.3cm}|p{2.3cm}|p{2.3cm}|p{2.3cm}|p{2.3cm}|p{3.4cm}|}
\hline
\multirow{2}{1.3cm}{\textbf{Pattern type}} & \multicolumn{4}{|c|}{\textbf{Graph pattern}} & \multirow{2}{*}{\textbf{Example response}} \\ \cline{2-5}
 & \textbf{Subject} & \textbf{Predicate} & \textbf{Object} & \textbf{Named Graph} &  \\ \hline \hline

\multicolumn{6}{|c|}{\textbf{Knowledge aspect: }\groupMetric{Correctness}} \\ \hline 
\multirow{6}{1.3cm}{Negation Conflict} & 
    \rdf{lWorld:<\underline{SUBJECT}>} & \rdf{n2mu:<\underline{PREDICATE}>} & \rdf{lWorld:<\underline{OBJECT}>} & \rdf{lTalk:<\underline{CLAIM}>}& 
    \multirow{6}{3.4cm}{"\textit{You say that Karla lives in Paris, but I have heard she does not}"} \\ 
 &  \rdf{lTalk:<MENTION1>} & \rdf{gaf:denotes} & \rdf{lTalk:<CLAIM>} & \rdf{lTalk:Perspectives}& \\ 
 &  \rdf{lTalk:<MENTION1>} & \rdf{grasp:hasAttribution} & \rdf{lTalk:<ATTRIBUTION1>} & \rdf{lTalk:Perspectives}& \\ 
 &  \rdf{lTalk:<ATTRIBUTION1>} & \rdf{rdf:value} & \rdf{graspf:Positive} & \rdf{lTalk:Perspectives}& \\ 
 &  \rdf{lTalk:<MENTION2>} & \rdf{gaf:denotes} & \rdf{lTalk:<CLAIM>} & \rdf{lTalk:Perspectives}&  \\ 
 &  \rdf{lTalk:<ATTRIBUTION2>} & \rdf{rdf:value} & \rdf{graspf:Negative} & \rdf{lTalk:Perspectives}& \\ \hline

\multirow{3}{1.3cm}{Cardinality Conflict} &  
    \rdf{n2mu:<PREDICATE>} & \rdf{owl:cardinality} & \rdf{"1"xsd:int} & \rdf{lWorld:Ontology}&
    \multirow{3}{3.4cm}{"\textit{I heard Karla lives in Amsterdam, not in Paris}"} \\
&   \rdf{lWorld:<\underline{SUBJECT}>} & \rdf{n2mu:<\underline{PREDICATE}>} & \rdf{lWorld:<\underline{OBJECT1}>} & \rdf{lWorld:<\underline{CLAIM1}>}& \\ 
&   \rdf{lWorld:<\underline{SUBJECT}>} & \rdf{n2mu:<\underline{PREDICATE}>} & \rdf{lWorld:<OBJECT2>} & \rdf{lWorld:<CLAIM2>}&  \\ \hline \hline

\multicolumn{6}{|c|}{\textbf{Knowledge aspect: }\groupMetric{Completeness}} \\ \hline 
\multirow{4}{1.3cm}{Subject Gap} & 
    \rdf{lWorld:<\underline{SUBJECT}>} & \rdf{n2mu:<\underline{PREDICATE}>} & \rdf{lWorld:<\underline{OBJECT}>} & \rdf{lTalk:<\underline{CLAIM}>}& 
    \multirow{4}{3.4cm}{"\textit{Karla is a person, and people are born in countries. Which country was Karla born in?}"} \\ 
&   \rdf{lWorld:<\underline{SUBJECT}>} & \rdf{rdf:type} & \rdf{n2mu:<TYPE1>} & \rdf{lWorld:Instances}&  \\ 
&   \rdf{n2mu:<\underline{PREDICATE}>} & \rdf{rdfs:domain} & \rdf{n2mu:<TYPE1>} & \rdf{lWorld:Ontology}& \\ 
&   \rdf{n2mu:<\underline{PREDICATE}>} & \rdf{rdfs:range} & \rdf{n2mu:<TYPE2>} & \rdf{lWorld:Ontology}& \\ \hline
    
\multirow{4}{1.3cm}{Object Gap} & 
    \rdf{lWorld:<\underline{SUBJECT}>} & \rdf{n2mu:<\underline{PREDICATE}>} & \rdf{lWorld:<\underline{OBJECT}>} & \rdf{lTalk:<\underline{CLAIM}>}&  
    \multirow{4}{3.4cm}{"\textit{Paris is a city, and cities are located in countries. Which country is Paris located in?}"} \\
&   \rdf{lWorld:<\underline{OBJECT}>} & \rdf{rdf:type} & \rdf{n2mu:<TYPE1>} & \rdf{lWorld:Instances}&   \\
&   \rdf{n2mu:<\underline{PREDICATE}>} & \rdf{rdfs:domain} & \rdf{n2mu:<TYPE1>} & \rdf{lWorld:Instances}& \\ 
&   \rdf{n2mu:<\underline{PREDICATE}>} & \rdf{rdfs:range} & \rdf{n2mu:<TYPE2>} & \rdf{lWorld:Instances}&  \\ \hline \hline

\multicolumn{6}{|c|}{\textbf{Knowledge aspect: }\groupMetric{Redundancy}} \\ \hline 
\multirow{3}{1.3cm}{Statement Novelty} & 
    \rdf{lWorld:<\underline{SUBJECT}>} & \rdf{n2mu:<\underline{PREDICATE}>} & \rdf{lWorld:<\underline{OBJECT}>} & \rdf{lTalk:<\underline{CLAIM}>}&  
    \multirow{3}{3.4cm}{"\textit{Gabriela also mentioned that Karla lives in Paris}"}  \\
&   \rdf{lTalk:<MENTION1>} & \rdf{gaf:denotes} & \rdf{lTalk:<\underline{CLAIM}>} & \rdf{lTalk:Perspectives}&  \\
&   \rdf{lTalk:<MENTION2>} & \rdf{gaf:denotes} & \rdf{lTalk:<\underline{CLAIM}>} & \rdf{lTalk:Perspectives}&  \\ \hline

\multirow{3}{1.3cm}{Entity Novelty} & 
    \rdf{lWorld:<\underline{SUBJECT}>} & \rdf{n2mu:<\underline{PREDICATE}>} & \rdf{lWorld:<\underline{OBJECT}>} & \rdf{lTalk:<\underline{CLAIM}>}&  
    \multirow{3}{3.4cm}{"\textit{I have heard many things about Paris}"} \\
&   \rdf{lWorld:<\underline{SUBJECT}>} & \rdf{grasp:denotedIn} & \rdf{lWorld:<MENTION1>} & \rdf{lTalk:Perspectives}&  \\
&   \rdf{lWorld:<\underline{SUBJECT}>} & \rdf{grasp:denotedIn} & \rdf{lWorld:<MENTION2>} & \rdf{lTalk:Perspectives}&  \\  \hline

\multicolumn{6}{|c|}{\textbf{Knowledge aspect: }\groupMetric{Interconnectedness}} \\ \hline 
\multirow{2}{1.3cm}{Subject Overlap} & 
    \rdf{lWorld:<\underline{SUBJECT}>} & \rdf{n2mu:<\underline{PREDICATE}>} & \rdf{lWorld:<\underline{OBJECT1}>} & \rdf{lTalk:<\underline{CLAIM1}>}&  
    \multirow{2}{3.4cm}{"\textit{You ate french food and now moroccan food.}" } \\
&   \rdf{lWorld:<\underline{SUBJECT}>} & \rdf{n2mu:<\underline{PREDICATE}>} & \rdf{lWorld:<OBJECT2>} & \rdf{lTalk:<CLAIM2>}& \\ \hline  

\multirow{2}{1.3cm}{Object Overlap} & 
    \rdf{lWorld:<\underline{SUBJECT1}>} & \rdf{n2mu:<\underline{PREDICATE}>} & \rdf{lWorld:<\underline{OBJECT}>} & \rdf{lTalk:<\underline{CLAIM1}>}&  
    \multirow{2}{3.4cm}{"\textit{My friend Armando also lives in Paris}" } \\
&   \rdf{lWorld:<SUBJECT2>} & \rdf{n2mu:<\underline{PREDICATE}>} & \rdf{lWorld:<\underline{OBJECT}>} & \rdf{lTalk:<CLAIM2>}& \\ \hline  
 
\end{tabular}}
\end{table*}

\begin{table*}[!ht]
\centering
\scriptsize{
\caption{
\label{table:ikg} Example of an interaction knowledge graph (iKG). The graph represents the interlocutor Marco, expressing the belief that "Diana lives in Paris", on January 14th, 2022. }
\begin{tabular}{|p{4cm}|p{3cm}|p{4cm}|p{3cm}|} 
\hline
\textbf{Subject} & \textbf{Predicate} & \textbf{Object} & \textbf{Named Graph} \\ \hline \hline
\rdf{lTalk:chat1\_turn1}    & \rdf{rdf:type}        & \rdf{grasp:Turn}      & \rdf{lTalk:Perspectives} \\
                            & \rdf{sem:hasActor}    & \rdf{lFriends:marco}  & \rdf{lTalk:Perspectives}\\
                            & \rdf{sem:hasTime}     & \rdf{lTime:14012022}  & \rdf{lTalk:Perspectives}\\
\hline
\rdf{lTalk:chat1\_turn1\_MEN1}  & \rdf{rdf:type}               & \rdf{grasp:Mention}                    & \rdf{lTalk:Perspectives} \\
                                & \rdf{grasp:denotes}          & \rdf{lWorld:diana\_live\_paris}        & \rdf{lTalk:Perspectives} \\
                                & \rdf{prov:wasDerivedFrom}    & \rdf{lTalk:chat1\_turn1}               & \rdf{lTalk:Perspectives} \\
                                & \rdf{grasp:hasAttribution}   & \rdf{lTalk:chat1\_turn1\_MEN1\_ATTR1}  & \rdf{lTalk:Perspectives} \\
\hline
\rdf{lTalk:chat1\_turn1\_MEN1\_ATTR1}   & \rdf{rdf:type}    & \rdf{grasp:Attribution}           & \rdf{lTalk:Perspectives} \\   
                                        & \rdf{rdf:value}   & \rdf{graspPolarity:positive}      & \rdf{lTalk:Perspectives} \\
                                        & \rdf{rdf:value}   & \rdf{graspCertainty:uncertain}    & \rdf{lTalk:Perspectives} \\
\hline
\end{tabular}}
\end{table*}





\begin{table*}[h]
\centering
\small{
\caption{
\label{table:users}User models and their knowledge communication qualities. User models with imperfect knowledge have their eKG corrupted, as these represent their belief networks.}
\begin{tabular}{lp{4.5cm}p{2.5cm}p{5cm}}
\hline
\multicolumn{4}{c}{\textbf{Perfect knowledge}} \\ \hline 
vanilla  & Oracle with perfect communication & NA & NA \\ \\

\hline
\multicolumn{4}{c}{\textbf{Imperfect knowledge}} \\ \hline 
amateur  & incomplete knowledge & coverage & 50\% claims removed \\
doubtful  & low confidence knowledge & certainty & 50\% claims with low certainty \\
incoherent  & conflicting knowledge & consistency & 50\% claims are negated \\ 
confused  & incorrect knowledge & correctness & 50\% claims with a random object \\
\hline
\end{tabular}}
\end{table*}

\end{document}